\documentclass[conference]{IEEEtran}
\IEEEoverridecommandlockouts
\usepackage{cite}
\usepackage{amsmath,amssymb,amsfonts}
\usepackage{algorithmic}
\usepackage{graphicx}
\usepackage{textcomp}
\usepackage{xcolor}

\usepackage{url}
\usepackage{amsthm}
\usepackage{booktabs}
\usepackage{algorithm}
\usepackage{algorithmic}
\usepackage{multirow}
\usepackage{xcolor}
\usepackage{subfigure}

\newcommand{\yankun}[1]{\textcolor{black}{#1}}

\def\BibTeX{{\rm B\kern-.05em{\sc i\kern-.025em b}\kern-.08em
    T\kern-.1667em\lower.7ex\hbox{E}\kern-.125emX}}
\begin{document}

\title{A Bidirectional Tree Tagging Scheme for Joint Medical Relation Extraction}

\author{\IEEEauthorblockN{1\textsuperscript{st} Xukun Luo}
\IEEEauthorblockA{
\textit{Peking University}\\
Beijing, China \\
luoxukun@pku.edu.cn}
\and
\IEEEauthorblockN{2\textsuperscript{nd} Weijie Liu}
\IEEEauthorblockA{
\textit{Peking University}\\
Beijing, China \\
dataliu@pku.edu.cn}
\and
\IEEEauthorblockN{3\textsuperscript{rd} Meng Ma}
\IEEEauthorblockA{
\textit{Peking University}\\
Beijing, China \\
mameng@pku.edu.cn}
\and
\IEEEauthorblockN{4\textsuperscript{th} Ping Wang}
\IEEEauthorblockA{
\textit{Peking University}\\
Beijing, China \\
pwang@pku.edu.cn}
}

\maketitle

\begin{abstract}
    Joint medical relation extraction refers to extracting triples, composed of entities and relations, from the medical text with a single model. One of the solutions is to convert this task into a sequential tagging task. However, in the existing works, the methods of representing and tagging the triples in a linear way failed to the overlapping triples, and the methods of organizing the triples as a graph faced the challenge of large computational effort. In this paper, inspired by the tree-like relation structures in the medical text, we propose a novel scheme called Bidirectional Tree Tagging (BiTT) to form the medical relation triples into two two binary trees and convert the trees into a word-level tags sequence. Based on BiTT scheme, we develop a joint relation extraction model to predict the BiTT tags and further extract medical triples efficiently. Our model outperforms the best baselines by 2.0\% and 2.5\% in F1 score on two medical datasets. What's more, the models with our BiTT scheme also obtain promising results in three public datasets of other domains.
\end{abstract}

\section{Introduction}

Medical relation extraction is an important task for building knowledge graphs in the medical domain. Relation extraction (RE) refers to extracting the relations between entity pairs contained in unstructured text such as electronic health records (EHRs) and medication package inserts (MPIs).


The early studies on RE are mainly fallen on pipeline methods, i.e., first identify the entities in a sequence by the named entity recognition (NER) module, and then classify the relation for each entity pair by the relation classification (RC) module \cite{Socher2012Semantic,Zeng2014Relation}. However, in the above methods, the error in NER module would be introduced into the subsequent RC module. Thus, the joint methods were put forward to solve this error propagation problem. These studies
can be roughly divided into several paradigms \cite{Miwa2016End,Zheng2017Joint,Zeng2018Extracting}, among them, one stream of work is to convert the RE task into the sequential tagging task.
\begin{figure}[t]
    \centering
    \includegraphics[width=\columnwidth]{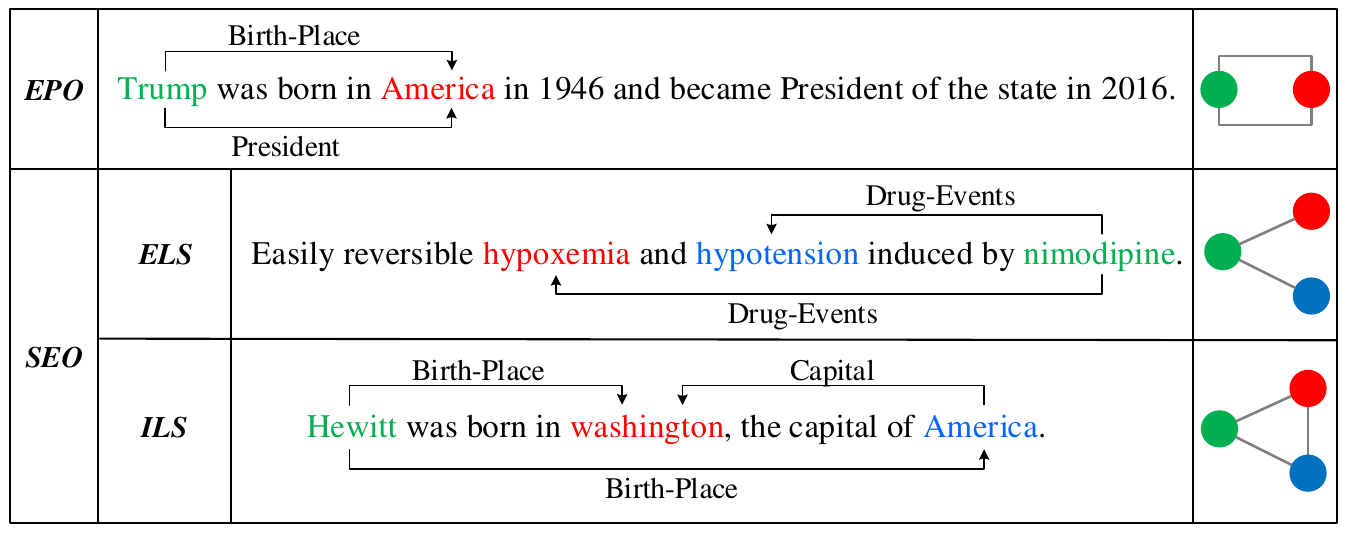}
    \caption{Examples of the sentences with overlapping triples, including \emph{EPO}, \emph{ELS} and \emph{ILS}.}
    \label{Fig:overlapping}
\end{figure}
The first work on tagging scheme \cite{Zheng2017Joint} is developed to organize the relation triples in a linear way and turn them into a tagging sequence with the same length $s$ as the input sentence. \yankun{As }\cite{Zheng2017Joint} needs to compute the probability of different tags for each token, 
the computational complexity is $O(s|\mathcal{R}|)$, where $|\mathcal{R}|$ is the size of the predefined relation set $\mathcal{R}$. However, 
\yankun{\cite{Zheng2017Joint}} failed to the overlapping triples, i.e., \emph{EntityPairOverlap (EPO)} and \emph{SingleEntityOverlap (SEO)} \cite{Zeng2018Extracting}. Take the first sentence in Figure \ref{Fig:overlapping} as an example, the tagging scheme of \cite{Zheng2017Joint} is unable to represent both \emph{Birth-Place} and \emph{President} in a single tag for each word. After that, some tagging schemes \cite{Dai2019Joint,Wang2020TPLinker} are proposed to handle the overlapping problem. They formed the relation triples into a graph represented by an adjacency matrix, and obtained a tagging sequence of length $s^2$. Though the improved tagging sequences were able to accommodate more information, the computation complexity of training 
\yankun{is increased from $O(s|\mathcal{R}|)$ }to $O(s^2|\mathcal{R}|)$.

\yankun{Besides, m}ost of 
existing medical RE frameworks are based on the RE models 
\yankun{specialized}
to the generic domains. However, the relation structure in the medical domain has different characteristics from other domains. \yankun{Specifically}, 
\yankun{from }EHRs and MPIs, we found that there are many tree-like relation structures in the medical text. For example, in the second sentence of Figure \ref{Fig:overlapping}, a drug (\emph{nimodipine}) corresponds to two effects (\emph{hypoxemia} and \emph{hypotension}). 
\yankun{Furthermore}
, drug contains multiple ingredients and can treat multiple symptoms in a MPI. 

Inspired by this discovery, in 
\yankun{our }paper, we employ a forest rather than a graph to represent relation triples in the medical sentence, and convert the forest into a tagging sequence of length $8s$. Specifically, our scheme called Bidirectional Tree Tagging (BiTT) is divided into three steps. First, the triples with the same relation in a sentence are grouped together. Second, the entities and relations in a group are modeled into two binary trees according to the order in which they appear in the sentence. Finally, we establish a mapping between the binary tree and token-level sequence tags so that they can be converted to each other. Our BiTT tags can represent more triples with low computation complexity of $O(s|\mathcal{R}|)$.

The key contributions are summarized as:

\begin{itemize}
    \item We propose a novel tagging scheme called BiTT, which apply the forest structure to represent the relation triples in the medical text;
	\item Based on our BiTT scheme, a joint medical RE model is developed for automatically predicting BiTT tags and extract relation triples;
	\item The models based on BiTT achieve solid results on two medical datasets and  three public generic datasets, which is a unique benefit obtained by effectively handling the overlap issue through the BiTT scheme.
\end{itemize}
\section{Related Work}
RE is a core task for text mining and information extraction. Depending on the structural diversity of neural networks, a number of pipeline RE models \cite{Zeng2014Relation,Xu2015Classifying,Vu2016Combining} were proposed with improved effect. The joint RE methods were put forward to solve the error propagation problem of the pipeline methods. Some works \cite{Miwa2016End,Katiyar2016Investigating} extended the RC module to share the encoder representation with the NER module, mitigated the problem of error propagation but still extracted entities and relations in a pipeline way. Some works \cite{Zheng2017Joint,Dai2019Joint,Wei2020CasRel} studied on the sequential tagging based methods, which turned the extracting tasks into a sequence labeling problem, bridging the information gap between the RC and NER steps. They focused on solving the overlapping triple problem and made great progress. However, these works could not deal with some special cases yet or cost much time. Some works \cite{Zeng2018Extracting,Nayak2019ptrnetdecoding,Zeng2020copymtl} directly generated the relational triples by the seq2seq framework. The seq2seq based works did not require complex tagging schemes, yet the maximum length of generated sequence was hard to determine and the decoding step cost much time.

Medical RE is a sub-field of RE, and researchers usually adapted the generic RE models through the challenges in the medical domain. For example, \cite{Li2016Joint} integrated the entity-related information, such as part-of-speech (POS) and medical ontology, into a joint RE model to improve the performance of medical entity recognition. \cite{Li2017A} proposed the trainable character embedding in the model to solve the out-of-vocabulary (OOV) problem in the medical text. And \cite{Qi2021KeMRE} exploited the medical knowledge of medicines, and incorporated the medical knowledge database into a RE model for Chinese medicine instructions.

\section{Methodology}

\begin{figure*}[t]
    \centering
    \includegraphics[width=.80\textwidth]{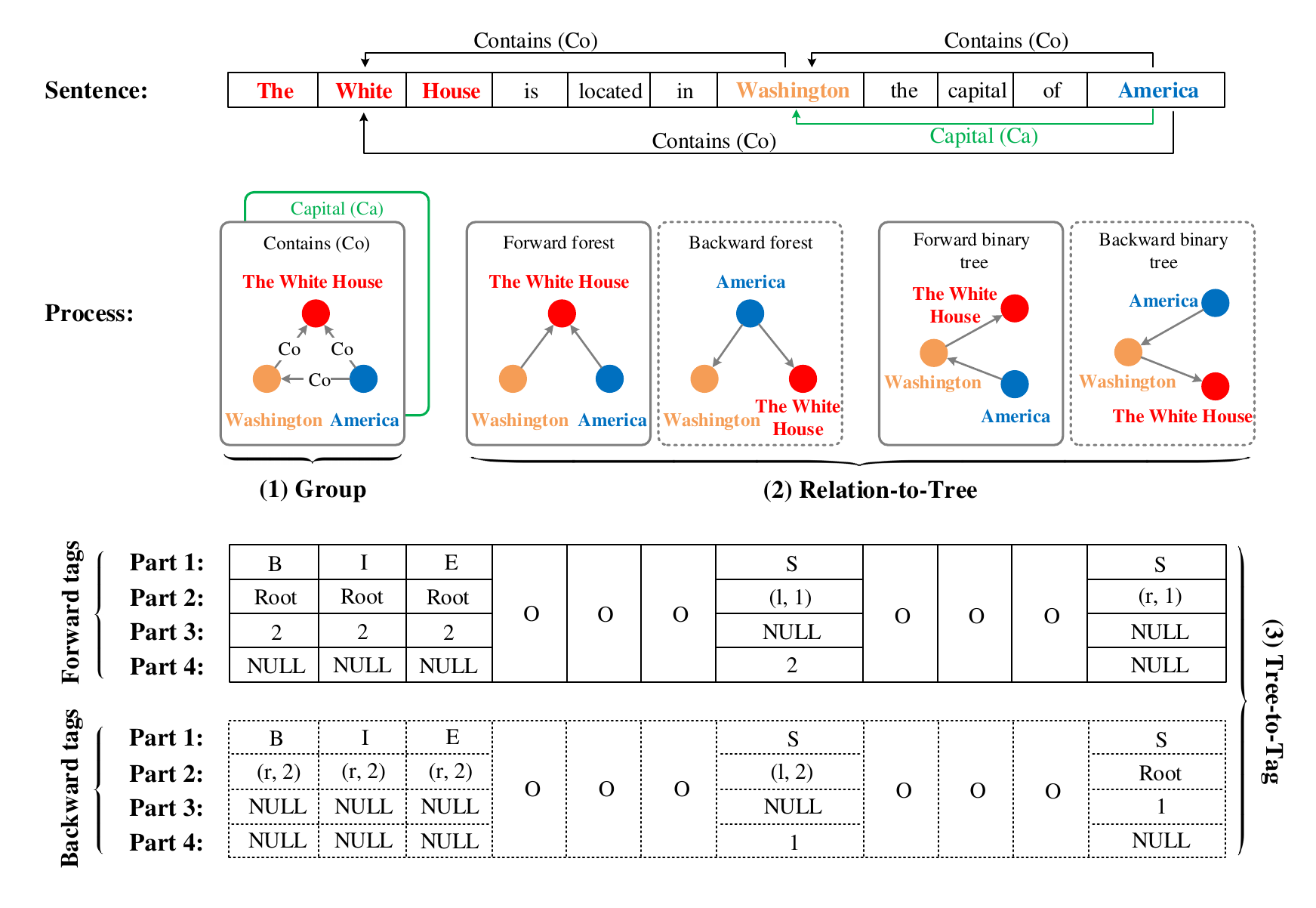}\\
    \caption{Our Bidirectional Tree Tagging (BiTT) Scheme. Take the triples with the relation category \emph{Contains} as examples.}
    \label{Fig:RelationsTreeTags}
\end{figure*}

In this section, we firstly present our fine-grained division for the sentences with overlapping triples. Then we illustrate how to convert these sentences to BiTT tag sequences and extract triples from the BiTT tags. Finally, we introduce our joint RE model for predicting BiTT tags.

\begin{algorithm}[tb]
    \caption{\emph{Relation-to-Tree}}
    \label{alg:turn}
    \begin{algorithmic}[1]
        \REQUIRE
            A $s$-words sentence, $S$;
            An array of $m$ relational triples with the same relation category in $S$, $RT$;
            An array of $n$ entities in $S$, $EN$;
        \ENSURE
            A binary relation tree, $B$;
        \STATE /* Construct a relation forest. */
        \STATE Initialize $L = []$, $F = []$;
        \FOR{each $i \in [1, n]$}
            \STATE Initialize $l$ = an array of all $EN_i$'s location pair in $S$;
            \STATE Add the elements in $l$ to $L$;
        \ENDFOR
        \STATE Sort $L$ from small to large according to beginning index;
        \WHILE {$L$ is not empty}
            \STATE Initialize a tree $T$ with the root $L_1$, remove $L_1$ from $L$;
            \FOR {$i = 2$; $i \leq length(L)$; $i ++$}
                \IF {$L_i$ not in $T$ \textbf{and} there is a valid relation between $L_i$ and the node in $T$}
                    \STATE Remove $L_i$ from $L$ and add it to $T$;
                \ENDIF
            \ENDFOR
            \STATE Add $T$ to $F$;
        \ENDWHILE
        \STATE /* Transform the forest to a binary tree. */
        \STATE Initialize an empty stack $S_t$, push the root nodes in $F$ to $S_t$ in order;
        \STATE Initialize a binary tree $B$ whose root node is $F_1$'s root;
        \FOR {$i = 2$; $i \leq length(F)$; $i ++$}
            \STATE Add $F_i$'s root to $B$ as the right child of $F_{i-1}$'s root;
        \ENDFOR
        \WHILE {$S_t$ is not empty}
            \STATE Initialize $node_{cur} = Pop(S_t)$, $C$ = the children array of $node_{cur}$ in $F$;
            \STATE Add $C_1$ to $B$ as the left child of $node_{cur}$;
            \FOR {$i = 2$; $i \leq length(C)$; $i ++$}
                \STATE Add $C_i$ to $B$ as the right child of $C_{i-1}$;
            \ENDFOR
            \STATE Push the children of $node_{cur}$ to $S_t$ in order;
        \ENDWHILE
    \end{algorithmic}
\end{algorithm}

\subsection{Fine-grained Division}
\label{ssec:Division}
As shown in Figure \ref{Fig:overlapping}, \cite{Zeng2018Extracting} categorized the sentences with overlapping triples into \emph{EPO} and \emph{SEO}. A sentence belongs to \emph{EPO} if there are some triples whose entity sets ($\{e_1, e_2\}$) are the same, and belongs to \emph{SEO} if there are some triples whose entity sets are different and contain at least one overlapping entity. Note that a sentence can pertain to the intersection of \emph{EPO} and \emph{SEO}. To separately handle the tree-like relation structures in the medical text, we further divide \emph{SEO} into \emph{ExcludeLoopSentences (ELS)} and \emph{IncludeLoopSentences (ILS)} based on the existence of relation loops. A sentence belongs to \emph{ILS} if there are some triples whose entity sets satisfy following conditions: (1) they are different; (2) each of them has at least one overlapping entity; (3) some of them contain two overlapping entities. A sentence belongs to \emph{ELS} if it pertains to \emph{SEO} but is not an \emph{ILS} sentence. Note that there is at least one loop in the relation graph of an \emph{ILS} sentence and no loop in the relation graph of an \emph{ELS} sentence without considering the edges' direction. Based on the statistics of the datasets from different domains in Table \ref{Tab:Dataset}, most of the medical sentences with overlapping triples belong to \emph{ELS}.

\subsection{Bidirectional Tree Tagging Scheme}
\label{ssec:BiTTScheme}
We propose a novel tagging scheme called BiTT, which uses two binary tree structures to incorporate the three \yankun{kinds of} sentence 
in Section \ref{ssec:Division}. We show an example in Figure \ref{Fig:RelationsTreeTags} to detail our handling approach and algorithm for different classes.

\subsubsection{EPO Handling}
\label{sssect:EPO}
Though there are some triples whose entity sets are the same in the \emph{EPO} sentences, their relation categories are different from each other. Hence we group 
the \yankun{corresponding} triples \yankun{together} with the same relation category in a sentence and label them respectively. For example, as shown in Figure \ref{Fig:RelationsTreeTags}, $(America, Capital, Washington)$ with the relation category \emph{Capital} is divided into an independent group, and 
triples with the relation category \emph{Contains} are aggregated into another group. Note that although the triple groups are labeled respectively, all triples are predicted simultaneously by the joint RE model to preserve their correlations.

\subsubsection{ELS Handling}
\label{sssect:ELS}
The most remarkable feature for the triples in an \emph{ELS} sentence is that, there is no loop in its relation graph, thus the triples can be completely represented as a forest. We handle the \emph{ELS} sentences with a Tree Tagging scheme, which consists of two steps.

\textbf{Relation-to-Tree} The pseudo code is presented in \textbf{Algorithm 1}. First, obtain the positions of all entities in a sentence and construct a relation forest according to the entities' forward appearing order. Take the first entity as the root of the first tree $T_1$, and then recurrently go through other entities 
from left to right, add the entities having relations with $T_1$'s nodes to $T_1$ gradually. When the remaining entities can no longer be added to the current tree, select the first entity of remains as a new tree $T_i$'s root and apply previous operations on $T_i$ until there is no entity left. After that, all trees are aggregated into a forest $F$. Second, we convert $F$ into a binary tree $B$. For an entity node $e$ in $F$, $e$'s first child becomes $e$'s left child in $B$, and right adjacent brother becomes $e$'s right child in $B$. For example, the node \emph{Washington} is the first child of \emph{The White House} and the left brother of \emph{America} in the forward forest in Figure~\ref{Fig:RelationsTreeTags}. Therefore, \emph{Washington} becomes the left child of \emph{The White House} and \emph{America} becomes \emph{Washington}'s right child in the forward binary tree. Note that we add the annotation $Brother$ on an edge in $B$ if the nodes connected by the edge used to be two root nodes in $F$. Besides, every edge is directional, from $e_1$ to $e_2$ in a specific triple.

\textbf{Tree-to-Tag} We assign a tag for every single word in the sentence according to the binary relation tree $B$ from \emph{Relation-to-Tree} step. If a word does not belong to any entity, its tag will be ``$O$". If a word is a part of an entity node $e$, its tag will be a combination of following four parts:
\begin{itemize}
	\item Part 1 ($P_1$) indicates the position of a word in the entity node $e$ with the 
    ``$BIES$" signs, i.e., $P_1 \in \{B(begin), I(in), E(end), S(single)\}$.

	\item Part 2 ($P_2$) indicates the information on the edge between $e$ and its parent in 
    $B$. $P_2 = Root$ when $e$ is the root of $B$. And $P_2 = Brother$ when the annotation on 
    the edge is also $Brother$. Except, $P_2 = (Child_2, Role_2)$, where 
    $Child_2 \in \{l(left), r(right)\}$ indicates if $e$ is the left or right child of its 
    parent, and $Role_2 \in \{1(e_1), 2(e_2)\}$ shows the entity role of $e$ pointed out by 
    the direction of the edge.

	\item Part 3 ($P_3$) indicates the information on the edge between $e$ and its left child 
    in $B$. $P_3 = NULL$ when $e$ has no left child. Except that, $P_3 = Role_3$, where 
    $Role_3 \in \{1(e_1), 2(e_2)\}$.

	\item Part 4 ($P_4$) indicates the information on the edge between $e$ and its right child 
    in $B$. $P_4 = NULL$ when $e$ has no right child. $P_4 = Brother$ when the annotation on 
    the edge is also $Brother$. Except, $P_4 = Role_4$, where $Role_4 \in \{1(e_1), 2(e_2)\}$.
\end{itemize}

\subsubsection{ILS Handling}
Since there is at least one loop in the relation graph of an \emph{ILS} sentence, the triples can not be absolutely represented by a single forest. To solve the problem, we simply build another forest according to the entities' backward appearing order, then convert it to a backward binary tree, and obtain the backward tree tags for a sentence. We apply the forward tags and the backward tags to accommodate more triples.

\subsection{Tags to Triples}
For the forward tags, first find all root nodes in the forest by the condition of $P_2 = Root$ or $P_2 = Brother$. Then we start from these root nodes, recursively match $P_3$ or $P_4$ of the nodes already in forest with other nodes' $P_2$ to reconnect the edges between the node pairs. If a parent node can match more than one node as its left (or right) child, we select the nearest node behind of it, since we take the entities’ appearing order into account when building the forest in the first step of the Tree Tagging scheme in Section~\ref{sssect:ELS}. What’s more, when rebuilding the relation tree, if a child or brother node is missed, our algorithm simply ignore it and discard its sub-tree. After reconstructing the forest, we can recursively extract the relational triples in the sentence from it.

The operation on the backward tags is the same as that on the forward tags. Except that when a parent node can match more than one node as its left (or right) child, we select the nearest node before it.

\subsection{Joint Relation Extraction Model}
\label{ssec:model}


In this section, we introduce our joint RE model in four parts, i.e., the text embedding module, the encoder module, the decoder module and our loss function respectively.

\subsubsection{Text Embedding}
For a given word $w$ in the input sentence, the representation $\boldsymbol{e} \in \mathbb{R}^{d}$ of $w$ from the text embedding module is concatenate by four parts:
\begin{equation}\label{equa:Embedding}
    \boldsymbol{e} = Linear([\boldsymbol{e}^w; \boldsymbol{e}^l; \boldsymbol{e}^c; \boldsymbol{e}^p])
\end{equation}
where $\boldsymbol{e}^w \in \mathbb{R}^{d_w}$ is the word embedding, $\boldsymbol{e}^l \in \mathbb{R}^{d_l}$ is the pre-trained contextualized word embedding from language models such as BERT \cite{Devlin2019BERT}, $\boldsymbol{e}^c \in \mathbb{R}^{d_c}$ is the character embedding, and $\boldsymbol{e}^p \in \mathbb{R}^{d_p}$ is the part-of-speech (POS) embedding.

\subsubsection{Encoder}
The encoder module aims to extract a context vector representation for each word in the sentence. In this paper, we adopt the Bi-directional Long Short-Term Memory (Bi-LSTM) layers and the multi-head self-attention layers as the encoder.

A Bi-LSTM layer consists of a forward LSTM \cite{Hochreiter1997Long} and a backward one. Denote the word embedding vectors of a sentence with $s$ words as $V = [\boldsymbol{e}_1, \ldots, \boldsymbol{e}_s]$. The output $\boldsymbol{o}_t$ and the hidden state $\boldsymbol{h}_t$ of the $t$-th word $\boldsymbol{w}_t$ from LSTM are:
\begin{equation}\label{equa:LSTM}
    \boldsymbol{o}_t, \boldsymbol{h}_t = lstm\_block(\boldsymbol{e}_t, \boldsymbol{h}_{t-1})
\end{equation}
where $lstm\_block (\cdot)$ is the function of a memory block in LSTM. We concatenate the two hidden states corresponding to the same word together as the Bi-LSTM hidden state. Thus, the Bi-LSTM hidden state $\dot{\boldsymbol{h}}_t$ of the $t$-th word $w_t$ is:
\begin{equation}\label{equa:BiLSTM}
    \dot{\boldsymbol{h}}_t = [\overrightarrow{\boldsymbol{h}_t}, \overleftarrow{\boldsymbol{h}_{s-t+1}}]
\end{equation}
where $\overrightarrow{\boldsymbol{h}_t}$ and $\overleftarrow{\boldsymbol{h}_{s-t+1}}$ are respectively the hidden states of $w_t$ in the forward LSTM and the backward one. 

The multi-head self-attention layer, the main component in the Transformer \cite{Vaswani2017Attention} encoder, is represented mathematically as follow:
\begin{align}\label{euqa:multiattention}
  MultiHead(\mathbf{H}) &= [head_1(\mathbf{H}); \ldots; head_h(\mathbf{H})] \\\label{equa:head}
  head_i(\mathbf{H}) &= Atten(\mathbf{H}\mathbf{W}^{Q}_i, \mathbf{H}\mathbf{W}^{K}_i, \mathbf{H}\mathbf{V}^{Q}_i) \\\label{equal:attention}
  Atten(\mathbf{Q}, \mathbf{K}, \mathbf{V}) &= softmax(\frac{\mathbf{Q}\mathbf{K}^T}{\sqrt{d_k}})\mathbf{V}
\end{align}
Here $\mathbf{H} \in \mathbb{R}^{s \times d_h}$ denotes the hidden state matrix $[\dot{\boldsymbol{h}}_1, \ldots, \dot{\boldsymbol{h}}_s]$ of the input sentence from the last Bi-LSTM layer. $\mathbf{W}^{Q}_i$, $\mathbf{W}^{K}_i$ and $\mathbf{V}^{Q}_i$ are learnable matrices. $h$ indicates the number of heads and $d_k = d_h / h$ means the size of the key vector.

\subsubsection{Decoder}
\label{sssec:decoder}
The decoder module is designed to parse the word representations and then predict the BiTT tags for each word. In this paper, it consists of a series of Linear layers. And there are two alternative construction methods for the Linear layers to predict BiTT tags, i.e., one-head and multi-head. Take a forward tree tag as example, the former means that concatenate the four parts of the tag as one label and predict it by a single Linear layer, like the work in \cite{Zheng2017Joint}. The latter predicts these four parts separately with four Linear layers, requiring fewer parameters. In this study, we employ the multi-head mechanism to reduce the computational costs.

\subsubsection{Loss Function}
The loss function for training the forward or backward tags is a weighted bias objective function defined by:
\begin{equation}\label{equa:loss}
    L = {\lambda_1}{L_1} + {\lambda_2}{L_2} + {\lambda_3}{L_3} + {\lambda_4}{L_4}
\end{equation}
where $L_j(j \in \{1, 2, 3, 4\})$ are the same bias cross entropy functions as \cite{Zheng2017Joint} for the four parts of BiTT tags respectively, $\lambda_j$ are the weights for $L_j$. We obtain $L^f$ for the forward tags and $L^b$ for the backward tags by Eq.\eqref{equa:loss}, then define the total loss $L^T$ of our framework as follow:
\begin{equation}\label{equa:totalloss}
    L^T = L^f + {\gamma}{L^b}
\end{equation}
where $\gamma$ is a weight hyperparameter.

\section{Experiments}
\label{sect:exper}
\subsection{Experimental Setting}
\textbf{Datasets}
We evaluate our BiTT scheme on five datasets, including two medical datasets (ADE \cite{Gurulingappa2012Development} and CMeIE \cite{Guan2020CMeIE}) and three datasets (NYT \cite{Riedel2010Modeling}, WebNLG \cite{Gardent2017Creating} and DuIE \cite{Li2019A}) from other domains. 

ADE, collected from the English medical reports, contains 4.2k samples with the relation \emph{Adverse-Effect}. Following previous work \cite{Yan2021A}, the samples with overlapping entities are filtered out and the 10-fold cross validation is performed for ADE in our experiment. CMeIE is a Chinese medical dataset constructed by several rounds of manual annotation. It consists of 28k sentences and 44 relations derived from medical textbooks and clinical practices. 

NYT contains 66.1k sentences with 24 relation categories from New York Times news. We adopt the preprocessed version released by \cite{Zeng2018Extracting}. WebNLG is an English dataset created for Natural Language Generation (NLG) task. It was adapted for RE task and released by \cite{Zeng2018Extracting}, which contains 6.2k sentences and 246 relation types. DuIE, consisting of 214.7k sentences with 49 relation categories, is a big Chinese dataset released by Baidu Inc. for RE. Since the testing set of DuIE is not available, we randomly separate the training set into a new validation set (20k samples) and a new training set (153.1k samples), and take the old validation set as the testing set (21.6k samples).

Table \ref{Tab:Dataset} illustrates some statistics of the samples with overlapping triples in the five datasets. It indicates that the overlapping triple problem is very common in all datasets. Besides, in the medical datasets, the samples with overlapping triples have a high proportion of \emph{ELS} samples , reaching more than 87\%. However, this pattern may not apply to the datasets in other domains, e.g., the sentences with overlapping triples have a low proportion of \emph{ELS} sentences at only 42.2\%.

\textbf{Metrics}
We employ micro Precision (\emph{Prec}), Recall (\emph{Rec}) and F1 score (\emph{F1}) to evaluate the performance of our models. We consider a predicted triple $(e_1, r, e_2)$ as a correct one only if $e_1$, $e_2$ and $r$ are all correct. Note that some original achievements of baselines evaluate the performance by Partial Matching, which means that $(e_1, r, e_2)$ is simply regarded as a correct triple if the first (or last) tokens of two entities ($e_1$ and $e_2$) and $r$ are correct.

\textbf{\yankun{Implementation} Details}
For the medical datasets, we apply the model called BiTT-Med mentioned in Section \ref{ssec:model} for prediction of BiTT tagging sequence and extraction of relation triples. In the text embedding module, the hidden sizes of different embedding vectors are $d_w = d_c = d_p = 100$ and $d_l = d = 768$. The pre-trained GloVe \cite{Pennington2014Glove} vectors are used to initialize the word embedding $e_w$. The contextualized word embedding $e_l$ is fixed and initialized by the vectors from the pre-trained BERT-base encoder. The character embedding is randomly initialized and computed by an LSTM \cite{Lample2016Neural} to cope with the OOV problem. And the POS tagging sequence of the input sentence is generated by SpaCy \cite{Honnibal2015An}. In the encoder module, the number of Bi-LSTM layers in the encoder module is 2 and the hidden size is 768. The number of multi-head self-attention layers is 2 and the number of head is $h = 8$. As for the loss function, in order to balance the loss of each part of the BiTT tags, we adopt $\lambda_1 = 8/6 $, $\lambda_2 = 8/8$, $\lambda_3 = 8/5$ and $\lambda_4 = 8/6$ in Eq.\eqref{equa:loss}, since the number of optional labels in the four parts are 6, 8, 5 and 6 after adding the tag ``O" and the padding tag respectively. And we set $\gamma = 1$ in Eq.\eqref{equa:totalloss}. Besides, the maximum length of an input sentence, the batch size and the learning rate are set to 128, 8 and 1e-4 respectively. And the training operation is terminated when the \emph{F1} on validation set no longer rises. Besides, in Table \ref{Tab:MedResult}, some baselines on CMeIE dataset is upgraded by ourselves for a further comparison. We simply substitute our text embedding module for the word embedding layers of the original achievements their papers.

\begin{table}[t]
    \centering
    \caption{Statistics of the training set and the testing set of the five datasets. \emph{ELS Ratio} indicates \emph{ELS} / Overlap Samples.}\label{Tab:Dataset}
    \resizebox{\columnwidth}{!}{
    \begin{tabular}{@{}lccccc@{}}
    \toprule
    Dataset & \textit{EPO} & \textit{ELS} & \textit{ILS} & \begin{tabular}[c]{@{}c@{}}Overlap\\ Samples\end{tabular} & \textit{\begin{tabular}[c]{@{}c@{}}ELS\\ Ratio\end{tabular}} \\ \midrule
    ADE \cite{Gurulingappa2012Development}     & 118          & 1,216        & 159          & 1,391                                                     & 0.874                                                        \\
    CMeIE \cite{Guan2020CMeIE}   & 381          & 8,805        & 457          & 9,213                                                     & 0.956                                                        \\ \midrule
    NYT \cite{Riedel2010Modeling}     & 17,004       & 10,740       & 2,006        & 25,422                                                    & 0.422                                                        \\
    WebNLG \cite{Gardent2017Creating}  & 622          & 2,894        & 1,294        & 3,957                                                     & 0.731                                                        \\
    DuIE \cite{Li2019A}    & 15,672       & 94,891       & 11,780       & 109,675                                                   & 0.865                                                        \\ \bottomrule
    \end{tabular}}
\end{table}

For the other three datasets, due to the large amount of training data and the absence of medical proper nouns in the generic datasets, we simplify BiTT-Med and obtain two new models called BiTT-LSTM and BiTT-BERT. Through these two models we can evaluate the performance of the BiTT scheme on generic domains. Specifically, we replace the text embedding layer of BiTT-Med with a learnable GloVe embedding in BiTT-LSTM and an unfixed BERT-base encoder in BiTT-BERT. The warm up rate in BiTT-BERT is set to 0.1. And we remove the multi-head self-attention layers of BiTT-Med in these two models. What's more, similar to the experiments on the medical datasets, in Table \ref{Tab:CommonResult}, we apply the BERT-base encoder as the word embedding layer of some baseline models and report the results. Note that the evaluation metrics in original NovelTagging and GraphRel are much looser than that for their upgraded version with BERT in our experiment, which makes the results of upgraded models with BERT look worse.

\begin{table}[t]
    \centering
    \caption{The performance comparison of different methods on ADE and CMeIE. $\heartsuit$ indicates that the results on CMeIE is upgraded from original achievements by our text embedding module. The main encoders applied in different models: L = LSTM, L+C = LSTM + CNN, ALB = ALBERT-xxlarge-v1, Bb = BERT-base, Bb+G = BERT-base + GCN.}\label{Tab:MedResult}
    \resizebox{\columnwidth}{!}{
        \begin{tabular}{@{}llllclclc@{}}
        \toprule
        Model                                                           &  & Encoder &  & \textit{Prec} &  & \textit{Rec}  &  & \textit{F1}   \\ \midrule
        \textbf{ADE}                                                    &  &         &  &               &  &               &  &               \\
        Neural Joint \cite{Li2016Joint}                &  & L       &  & 64.0          &  & 62.9          &  & 63.4          \\
        Multi-head \cite{Bekoulis2018Joint}            &  & L       &  & 72.1          &  & 77.2          &  & 74.5          \\
        Multi-head + AT \cite{Bekoulis2018Adversarial} &  & L       &  & -             &  & -             &  & 75.5          \\
        Rel-Metric \cite{Tran2019Neural}               &  & L+C     &  & \underline{77.4}          &  & \underline{77.3}          &  & 77.3          \\
        Table-Sequence \cite{Wang2020Two}              &  & ALB     &  & -             &  & -             &  & \underline{80.1}          \\
        PFN \cite{Yan2021A}                            &  & Bb      &  & -             &  & -             &  & 80.0          \\ \midrule
        BiTT-Med (Ours)                                                &  & Bb      &  & \textbf{83.1} &  & \textbf{81.3} &  & \textbf{82.1} \\ \midrule
        \textbf{CMeIE}                                                    &  &         &  &               &  &               &  &               \\
        NovelTagging \cite{Zheng2017Joint} \textsuperscript{$\heartsuit$}             &  & Bb       &  & 51.4             &  & 17.1             &  & 25.6             \\
        GraphRel-1p \cite{Fu2019Graphrel} \textsuperscript{$\heartsuit$}              &  & Bb+G     &  & 31.2             &  & 26.0             &  & 28.4             \\
        GraphRel-2p \cite{Fu2019Graphrel} \textsuperscript{$\heartsuit$}              &  & Bb+G     &  & 28.5             &  & 23.1             &  & 25.5             \\
        CasRel \cite{Wei2020CasRel} \textsuperscript{$\heartsuit$}              &  & Bb     &  & \underline{53.5}             &  & \underline{28.2}             &  & 37.0             \\
        ER+RE \cite{Zhang2022CBLUE}              &  & ALB     &  & -             &  & -             &  & \underline{47.6}             \\ \midrule
        BiTT-Med (Ours)                                                &  & Bb      &  & \textbf{55.6}             &  & \textbf{45.5}             &  & \textbf{50.1}             \\ \bottomrule
        \end{tabular}}
\end{table}

\begin{table}[t]
    \centering
    \caption{The performance comparison of different methods on NYT, WebNLG and DuIE. $\spadesuit$ indicates the metrics for the models follow Partial Matching. $\heartsuit$ indicates that the results on NYT and DuIE is upgraded from original achievements by the BERT-base encoder. The main encoders applied in different models: L = LSTM, L+G = LSTM + GCN, Bb = BERT-base, Bb+G = BERT-base + GCN.}\label{Tab:CommonResult}
    \resizebox{\columnwidth}{!}{
    \begin{tabular}{@{}lllllllll@{}}
    \toprule
    Model            &  & Encoder &  & \textit{Prec} &  & \textit{Rec}  &  & \textit{F1}   \\ \midrule
    \textbf{NYT}     &  &         &  &               &  &               &  &               \\
    NovelTagging \cite{Zheng2017Joint} \textsuperscript{$\spadesuit$}     &  & L       &  & 62.4          &  & 31.7          &  & 42.0          \\
    CopyRE-Mul \cite{Zeng2018Extracting} \textsuperscript{$\spadesuit$}       &  & L       &  & 61.0          &  & 56.6          &  & 58.7          \\
    GraphRel-2p \cite{Fu2019Graphrel} \textsuperscript{$\spadesuit$}      &  & L+G     &  & 63.9          &  & 60.0          &  & 61.9          \\
    PA \cite{Dai2019Joint}               &  & L       &  & 49.4          &  & 59.1          &  & 53.8          \\
    CopyMTL-Mul \cite{Zeng2020copymtl}      &  & L       &  & 75.7          &  & 68.7          &  & 72.0          \\
    NovelTagging \cite{Zheng2017Joint} \textsuperscript{$\heartsuit$}     &  & Bb      &  & \underline{89.0}     &  & 55.6          &  & 69.3          \\
    CopyRE-Mul \cite{Zeng2018Extracting} \textsuperscript{$\heartsuit$}        &  & Bb &  & 39.1          &  & 36.5          &  & 37.8          \\
    GraphRel-2p \cite{Fu2019Graphrel} \textsuperscript{$\heartsuit$}      &  & Bb+G    &  & 82.5          &  & 57.9          &  & 68.1          \\
    CasRel \cite{Wei2020CasRel} \textsuperscript{$\spadesuit$}           &  & Bb      &  & \textbf{89.7} &  & \textbf{89.5} &  & \textbf{89.6} \\ \midrule
    BiTT-LSTM (Ours) &  & L       &  & 66.5          &  & 76.3          &  & 71.1          \\
    BiTT-BERT (Ours) &  & Bb      &  & \textbf{89.7} &  & \underline{88.0}    &  & \underline{88.9}    \\ \midrule
    \textbf{WebNLG}  &  &         &  &               &  &               &  &               \\
    NovelTagging \cite{Zheng2017Joint} \textsuperscript{$\spadesuit$}     &  & L       &  & 52.5          &  & 19.3          &  & 28.3          \\
    CopyRE-Mul \cite{Zeng2018Extracting} \textsuperscript{$\spadesuit$}       &  & L       &  & 37.7          &  & 36.4          &  & 37.1          \\
    GraphRel-2p \cite{Fu2019Graphrel} \textsuperscript{$\spadesuit$}      &  & L+G     &  & 44.7          &  & 41.1          &  & 42.9          \\
    CopyMTL-Mul \cite{Zeng2020copymtl}      &  & L       &  & 58.0          &  & 54.9          &  & 56.4          \\
    TPLinker \cite{Wang2020TPLinker}         &  & Bb      &  & \underline{88.9}    &  & \textbf{84.5} &  & \textbf{86.7} \\ \midrule
    BiTT-LSTM (Ours)       &  & L       &  & 83.8          &  & 66.0          &  & 73.8          \\
    BiTT-BERT (Ours) &  & Bb      &  & \textbf{89.1} &  & \underline{83.0}    &  & \underline{86.2}    \\ \midrule
    \textbf{DuIE}    &  &         &  &               &  &               &  &               \\
    NovelTagging \cite{Zheng2017Joint} \textsuperscript{$\heartsuit$}     &  & Bb      &  & \underline{75.0}    &  & 38.0          &  & 50.4          \\
    GraphRel-1p \cite{Fu2019Graphrel} \textsuperscript{$\heartsuit$}      &  & Bb+G    &  & 52.2          &  & 23.9          &  & 32.8          \\
    GraphRel-2p \cite{Fu2019Graphrel} \textsuperscript{$\heartsuit$}      &  & Bb+G    &  & 41.1          &  & 25.8          &  & 31.8          \\
    CaseRel \cite{Wei2020CasRel} \textsuperscript{$\heartsuit$}          &  & Bb      &  & \textbf{75.7} &  & \underline{80.0}    &  & \underline{77.8}    \\ \midrule
    BiTT-BERT (Ours) &  & Bb      &  & \textbf{75.7} &  & \textbf{80.6} &  & \textbf{78.0} \\ \bottomrule
    \end{tabular}}
\end{table}

\subsection{Baseline Models}
For comparison, we employ 13 models as baselines, which can be roughly divided into one-stage models and two-stage models. The one-stage models output both entities and relations at the same time, including Neural Joint \cite{Li2016Joint}, NovelTagging \cite{Zheng2017Joint}, Rel-Metric \cite{Tran2019Neural}, PA \cite{Dai2019Joint}, GraphRel \cite{Fu2019Graphrel}, Table-Sequence \cite{Wang2020Two}, TPLinker \cite{Wang2020TPLinker}, and PFN \cite{Yan2021A}. The two-stage models identify the entities firstly and then classify the relations of all entity pairs, or identify the head entities first and then find out the tail entities based on the relation categories for each head entity. \yankun{The corresponding representative methods include }Multi-head \cite{Bekoulis2018Joint}, Multi-head + AT \cite{Bekoulis2018Adversarial}, CopyRE \cite{Zeng2018Extracting}, CopyMTL \cite{Zeng2020copymtl} and CasRel \cite{Wei2020CasRel}. 

\subsection{Results}

\textbf{Compared Results on Medical Datasets} Table \ref{Tab:MedResult} presents the comparison of BiTT-Med with previous methods on two medical datasets. Our BiTT-Med achieves solid \emph{F1} scores on ADE and CMeIE, which are 82.1\% and 50.1\%. On ADE, our model outperforms Table-Sequence by 2.0\% in \emph{F1}, and outperforms Rel-Metric by 5.7\% in \emph{Prec} and 4.0\% in \emph{Rec}. On CMeIE, our model outperforms ER+RE by 2.5\% in \emph{F1}, and outperforms CasRel by 2.1\% in \emph{Prec} and 17.3\% in \emph{Rec}. This indicates the effectiveness of our model for jointly extracting medical entities and relations.

\textbf{Compare Results on Common Datasets} To further evaluate the performance of our BiTT scheme on the joint RE task, we apply our BiTT-LSTM and BiTT-BERT models on three generic datasets. Table \ref{Tab:CommonResult} shows the comparison of our models with baselines on NYT, WebNLG and DuIE.
For BiTT-LSTM, it achieves ideal \emph{F1} scores on NYT and WebNLG, which are 71.1\% and 73.8\%. BiTT-LSTM outperforms CopyMTL-One by 7.6\% and 11.1\% in \emph{Rec} on NYT and WebNLG. There is an impressive \emph{Rec} gap between BiTT-LSTM and the baselines without BERT encoder, which verifies the utility of BiTT scheme when dealing with overlapping triples. For BiTT-BERT, it also achieves solid \emph{F1} scores close to CasRel and TPLinker on NYT, WebNLG and DuIE datasets, which are 88.9\% , 86.2\% and 78.0\% respectively. And BiTT-BERT achieve the best \emph{Prec} on three datasets, which are 89.7\%, 89.1\% and 75.7\%. In addition, we note that the sequential tagging based models (NovelTagging, CasRel, TPLinker and BiTT-BERT) achieve higher \emph{Prec} than other models. It proves the superiority of the sequential tagging based methods for conservative prediction. However, BiTT-BERT does not perform as well as the best method on NYT and WebNLG. It probably results from the fact that the proportion of \emph{ELS} sentences in the samples with overlapping triples of NYT and WebNLG are 42.2\% and 73.1\%, which are not as large as in DuIE. This demonstrates the advantages of our BiTT scheme for handling \emph{ELS} sentences.

\subsection{Efficiency of BiTT based models} 
\begin{figure}[t]
    \centering
    \includegraphics[width=.80\columnwidth]{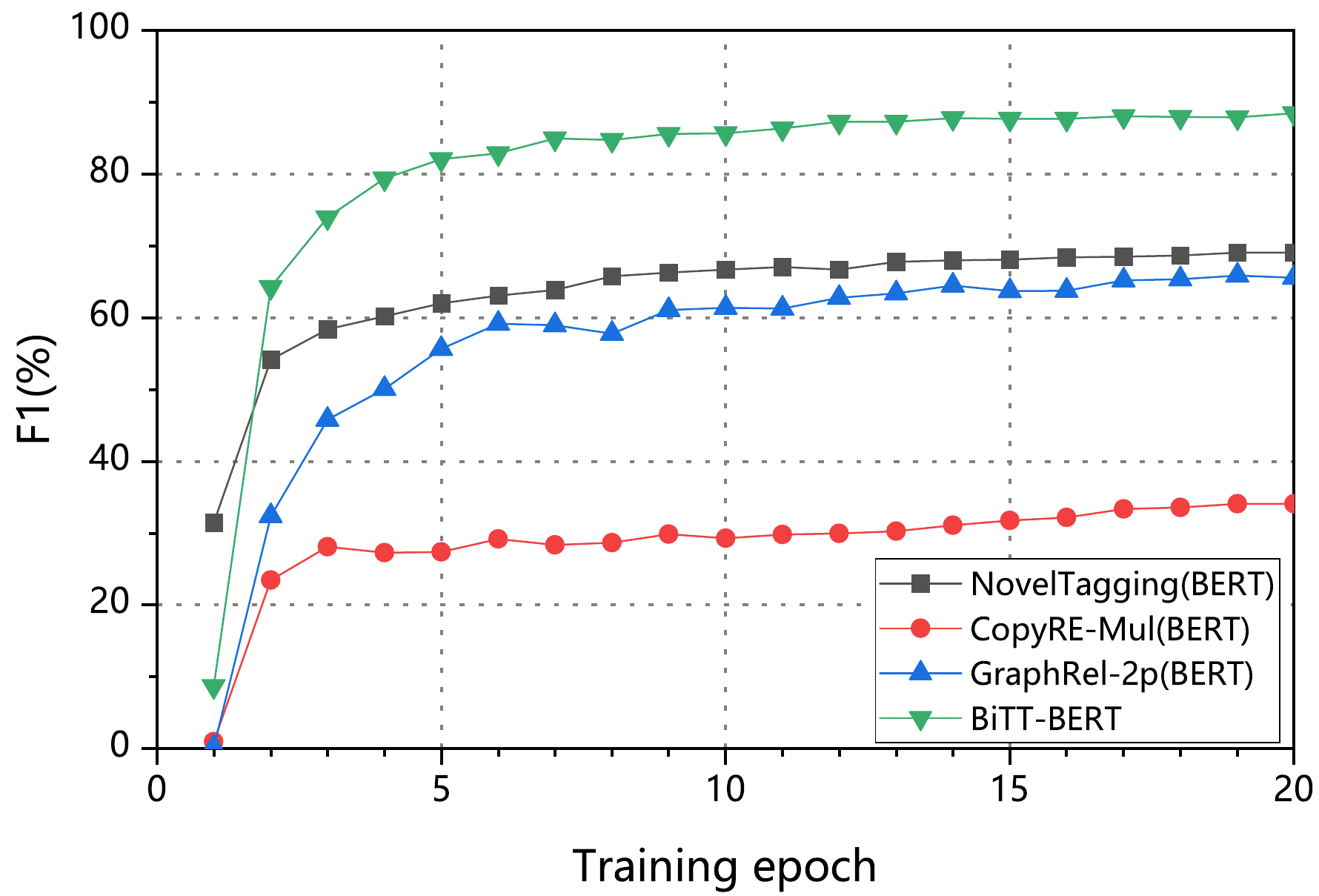}\\
    \caption{The training curves of different models with BERT-base encoder on the NYT dataset.}
    \label{Fig:Training}
\end{figure}

Our BiTT based models can predict the BiTT tags with low computation complexity of $O(s|\mathcal{R}|)$, thus have higher efficiency compared to graph-based and two-stage models. First, we compare our BiTT-BERT model with the one-stage baselines. As shown in Figure \ref{Fig:Training}, in our experiments on NYT, most of baseline models with BERT-base encoder converge at a similar epoch, i.e., about the 15th epoch. 
Besides, the number of decoder parameters in our BiTT-BERT (48.9M) is almost the same as that in NovelTagging (47.3M) and much less than that in other one-stage baseline models, i.e., GraphRel-2p (106.4M) and TPLinker (1293.3M). This indicates that with the same encoder, the converge speed of our framework is much faster than Graph-Rel and TPLinker.
Second, we compare \yankun{our} BiTT-BERT with the competitive two-stage baseline CasRel. We measure the time of a training epoch (traversing all triples in the WebNLG training set) for BiTT-BERT and CasRel on a same NIVIDIA GeForce RTX 2080 Ti. The result is that BiTT-BERT takes 255.0s and CasRel takes 1701.2s. CasRel takes about 6.7 times longer to traverse the dataset than BiTT-BERT, since the two-stage model CasRel copies a sentence into multiple samples for training based on the number of head entities.

\subsection{Ablation Study}
As is shown in Table \ref{Tab:Ablation}, to explore the effects of the handling methods for overlapping triples in BiTT scheme and our decoder architecture, we perform the ablation tests based on BiTT-BERT and the NYT dataset. BiTT-BERT ``w/o Group" means that the ``\emph{EPO} Handling" described in Section \ref{sssect:EPO} is dropped \yankun{while} replaced by adding the relation categories information to $P_2$, $P_3$ and $P_4$ of the BiTT tags. BiTT-BERT ``w/o Bidirectional" implies that we only build the forward forest in a sentence and then generate the forward tags, 
BiTT-BERT ``w/o Multi-head" indicates that the multi-head structure in our framework is substituted by the one-head structure, which is illustrated in the decoder module of our joint RE model in Section \ref{ssec:model}.

From Table \ref{Tab:Ablation}, we have observed some impressive facts when comparing these three ablation tests with BiTT-BERT. First, the grouping operation greatly improves the performance of our framework on not only \emph{EPO} (17.2\%) but also \emph{ELS} (11.1\%) and \emph{ILS} (17.2\%), since it reduces a complex relation graph into several simpler ones in a sentence. And it also reduces the categories of BiTT tags, since the information of relation categories and entity types does not need to be added to the tags. Second, the backward forest can effectively supplement the triples that cannot be totally represented by a single forward forest, which provides an \emph{F1} score boost of 10.8\% for \emph{ILS} sentences. The backward forest also boosts the \emph{F1} score on \emph{EPO} and \emph{ELS} sentences by 1.0\% and 2.8\%. This indicates that when the sub-tree of a node is discarded due to the wrong prediction of the forward forest, the backward forest can supplement the information of the missing node. Last, compared with the one-head structure, the multi-head structure can decrease the parameters of the decoder module, thus lessen the computation burden and the training time in an epoch. Besides, the F1 score of multi-head is 1.2\% higher than one-head. This indicates that multi-head does not make the interaction of the four parts of BiTT labels less, but makes the labels with fewer occurrences to be trained more.

\begin{table}[t]
    \centering
    \caption{The results of the ablation study on BiTT-BERT model and the NYT dataset. F1-EPO, F1-ELS, F1-ILS are the F1 scores on the EPO, ELS and ILS sentences. F1-All is the F1 score on all sentences. The scale of decoder parameters is Million (M), and the scale of training time is second per epoch (s/epoch).}\label{Tab:Ablation}
    \resizebox{\columnwidth}{!}{
    \begin{tabular}{@{}clcccc@{}}
    \toprule
    \multicolumn{2}{c}{Metrics}                        & \begin{tabular}[c]{@{}c@{}}w/o\\ Group\end{tabular} & \begin{tabular}[c]{@{}c@{}}w/o\\ Bidirectional\end{tabular} & \begin{tabular}[c]{@{}c@{}}w/o\\ Multi-head\end{tabular} & BiTT-BERT       \\ \midrule
    \multicolumn{2}{c}{\textit{F1-EPO}}            & 74.0                                                          & 90.2                                                                  & 90.2                                                               & \textbf{91.2}   \\
    \multicolumn{2}{c}{\textit{F1-ELS}}            & 76.2                                                          & 84.5                                                                  & 85.5                                                               & \textbf{87.3}   \\
    \multicolumn{2}{c}{\textit{F1-ILS}}            & 68.3                                                          & 74.7                                                                  & 81.8                                                               & \textbf{85.5}   \\
    \multicolumn{2}{c}{\textit{F1-All}}  & 81.9                                                          & 88.2                                                                  & 87.7                                                               & \textbf{88.9}   \\
    \multicolumn{2}{c}{Decoder Params} & 71.3                                                          & \textbf{48.0}                                                         & 68.5                                                               & 48.9            \\
    \multicolumn{2}{c}{Training Time}        & -                                                             & -                                                                     & 2125.0                                                             & \textbf{1739.0} \\ \bottomrule
    \end{tabular}}
\end{table}

\section{Conclusion}
In this paper, motivated by the tree-like relation structures in the medical text, we propose a Bidirectional Tree Tagging (BiTT) scheme to label the overlapping entities and relations in medical sentences with solid accuracy and high efficiency. We build up a jointly RE model called BiTT-Med and two simplified versions (BiTT-LSTM and BiTT-BERT) for experiments. The results on two public medical datasets and three generic datasets demonstrate that our proposal outperforms several baselines, especially when processing the \emph{ELS} cases.

Our future work aims to improve the BiTT scheme and the RE models \yankun{in following aspects}. (1) When extracting results from BiTT tags, we will try reconstructing a binary forest instead of a binary tree to reduce the error propagation if a node is dropped. (2) More rule restrictions for BiTT will be proposed to make it more robust. (3) We will apply more powerful pre-trained encoders to the extraction framework for better performance.

\end{document}